\documentclass[11pt,a4paper,UTF8]{article}
\usepackage{CJKutf8}
\usepackage[hyperref]{eacl2021}
\usepackage{times}
\usepackage{latexsym}
\usepackage{amsmath}
\usepackage{booktabs}

\usepackage{microtype}

\aclfinalcopy 

\usepackage{graphicx}
\usepackage{adjustbox}
\usepackage{enumitem}

\title{Don't Change Me! User-Controllable Selective Paraphrase Generation}

\author{
Mohan Zhang$^{1,2}$,
Luchen Tan$^1$,
Zhengkai Tu$^1$,
Zihang Fu$^1$,
Kun Xiong$^1$,
Ming Li$^{1,3}$,
Jimmy Lin$^{1,3}$
\\[1ex]
$^1$RSVP.ai, $^2$ University of Toronto, 
\\
$^3$David R. Cheriton School of Computer Science, University of Waterloo
\\[1ex]
\texttt{\small zhangmo4@cs.toronto.edu, lctan@rsvp.ai, siriuself1103@gmail.com,}
\\
\texttt{\small \{zhfu,kun\}@rsvp.ai, \{mli,jimmylin\}@uwaterloo.ca}
}

\date{}

\begin{document}
\maketitle
\begin{abstract}
In the paraphrase generation task, source sentences often contain phrases that should not be altered. Which phrases, however, can be context dependent and can vary by application.
Our solution to this challenge is to provide the user with explicit tags that can be placed around any arbitrary segment of text to mean ``don't change me!''\ when generating a paraphrase; the model learns to explicitly copy these phrases to the output.
The contribution of this work is a novel data generation technique using distant supervision that allows us to start with a pretrained sequence-to-sequence model and fine-tune a paraphrase generator that exhibits this behavior, allowing user-controllable paraphrase generation.
Additionally, we modify the loss during fine-tuning to explicitly encourage diversity in model output.
Our technique is language agnostic, and we report experiments in English and Chinese.
\end{abstract}

\section{Introduction}

Notions of semantic similarity and paraphrase are highly context dependent.
Consider ``I'm looking for cheap hotels in New York.'' vs.\ ``What are cheap lodging options in Beijing?'':\ from the perspective of intent classification, both express similar intents, but from the perspective of paraphrasing in a community QA application, a user looking for the answer to one question would not find the other response helpful.
This is because the location ``New York'' anchors the information need, and any changes to it would be unacceptable to the user.

It is not always the case that named entities are ``immutable'' in this respect:\ consider a user looking for vacation destinations in the South of France.
From the perspective of an advertiser, there might be good reason to tempt the user with alternative locations such as the Italian Riviera; the user may even welcome these suggestions.
Also, it is not always the case that these immutable anchors are named entities:\
For example, some metrics assign high similarity to antonyms, and so ``cheap hotels'' and ``expensive hotels'' might be considered semantically close, but obviously not from the perspective of an end user looking for inexpensive lodging.

Although whether or not certain words can be changed without affecting the meaning of a sentence is highly dependent on context, the user of a paraphrase generation system usually would know.
Consider the application of paraphrase generation in a community QA application, where a developer wishes to automatically generate question variants to increase the chances of a semantic match:\
A na\"ive system will indeed generate ``What are cheap lodging options in Beijing?''\ as a paraphrase to ``I'm looking for cheap hotels in New York.''

What if we provide the user with a way to explicitly tag portions of the input so that a paraphrase generator knows what parts of the input to repeat verbatim?
For example, a simple annotation scheme like ``What are cheap lodging options in $\langle$tag$\rangle$ Beijing $\langle/$tag$\rangle$?'', where words between $\langle$tag$\rangle$ and $\langle/$tag$\rangle$ should {\it not} be paraphrased.
We present a paraphrase generator that implements such tags, allowing user-controllable paraphrase generation.

On a standard sequence-to-sequence model (mBART),
our contribution is a novel data generation technique via distant supervision to fine-tune a paraphrase generator that supports this tagging behavior.
To our knowledge, we are the first to describe such a capability---this solves a practical problem that hinders deployment of models in real-world applications.\footnote{This is a feature requested by many of our customers.}
Our technique is entirely language agnostic, and we report experiments in both English and Chinese.
Additionally, we modify the loss during fine-tuning to explicitly encourage diversity in the paraphrase generation process.

\section{Paraphrase Generator}

We treat paraphrase generation as a standard supervised sequence-to-sequence task by fine-tuning mBART-large \citep{mbart}.
Although our technique can be applied to any sequence-to-sequence framework, we selected mBART because a multi-lingual model is widely available.
Following standard practice, custom language tags are used to denote the desired behavior of the model, but otherwise everything in our model is language agnostic.

Building on our running example, a training pair might be (``What are cheap lodging options in $\langle$tag$\rangle$ Beijing $\langle/$tag$\rangle$?'', ``I'm looking for cheap hotels in $\langle$tag$\rangle$ Beijing $\langle/$tag$\rangle$?'').
Multiple such corpora exists (without these tags), and this is a straightforward use of mBART, but the challenge is this:\ Starting with an existing paraphrase dataset, how do we know where the tags should go?
At inference time, the user supplies the tags, but during training, such knowledge is not available.
Obviously, we could go through and manually insert tags to existing paraphrase datasets, but such large annotation efforts are impractical.
Instead, we adopt a solution based on distant supervision by building three different taggers, discussed below.

\subsection{Taggers}

The {\bf Oracle Tagger} represents an upper bound:\ tags are assigned to surround consecutive word sequences that appear in both the source and multiple references.
Our intuition is that tags should go around exactly the portions of text that do not change across the paraphrases.
Here, we heuristically filter out stopwords and other common $n$-grams.
However, the Oracle Tagger won’t work on datasets like QQP, where each sentence only has one reference.
The two similar sentences are going to largely overlap with each other, so the Oracle Tagger will almost tag the entire sentence, which cannot be regarded as the anchors.
For datasets like MSCOCO, since there are multiple references, the overlapping  substrings are only a few words long therefore we can treat them as anchors.

With the {\bf NER Tagger}, we simply tag all NERs.
Since we aim to generate paraphrases in multiple languages, we use the ID-CNN language-independent named entity recognizer \citep{strubell2017fast}.
With the {\bf Auto Tagger}, we use the output of the oracle tags to train a standard BERT-based \citep{devlin2018bert} token classifier.

\subsection{Encouraging Diversity}

During fine-tuning, our paraphrase generator learns to keep the content between $\langle$tag$\rangle$ and $\langle/$tag$\rangle$ tokens---provided by one of the three taggers above---since the content inside are not changed from a source to its reference output.
We want our model to be able to keep the anchors (that are tagged) but paraphrase the other parts as much as possible.

To accomplish this, in addition to the original mBART architecture, we add another loss term to encourage diversity in generated paraphrases.
During fine-tuning, we also minimize the mutual information between our paraphrase distribution and the source sentence.
The mutual information term is controlled by a hyperparameter weight, indicating how ``different and diverse'' we want our paraphrases to be compared to the source sentence.
By default, this weight is set to 0.3.

To be more specific, let $|\mathcal{D}|$ be the number of tokens in our dictionary, $\epsilon$ be the weight of label smoothing, $w$ be the weight of the entropy term in our mutual information evaluated with source sentences.
Note that a larger $w$ means the more diversity we are encouraging our paraphrases to be.
Let $B$ be the batch size, $p(t|s_i)$ be the ground truth one-hot token probability distribution of reference sentences given a source sentence $s_i$ where $i \in \{1, 2, ..., B\}$, and let $q_{\theta}(t|s_i)$ be the model predicted token probability distribution given a source sentence $s_i$ for a fixed set of parameters $\theta$.
Our loss function can be written as:
\begin{equation} \label{eq1}
\begin{split}
\mathcal{L} = &  (1 - \epsilon) [- \sum_{i} p(\bf{t}|\bf{s_i})\log q_{\theta}(\bf{t}|\bf{s_i})]  \\
& + \sum_{i} [ - \frac{\epsilon}{|\mathcal{D}|} \log q_{\theta}(\bf{t}|\bf{s_i})] \\
& - w [- \sum_{i} p(\bf{t}==\bf{s_i})\log q_{\theta}(\bf{t}|\bf{s_i})]
\end{split}
\end{equation}
\noindent where $p(\bf{t}==\bf{s_i})$ can also be viewed as a vector of results of an indicator function; the value of such a vector is one at the $j$th location if and only if for a source sentence $s_i$ with length $J$, $t_j$ and $s_{i, j}$ are of the same token for $j \in {1,2,...,J}$.
\section{Experimental Setup}

We describe the data we used for experiments and the basic setup.
At a high level, we have two types of datasets, with or without paraphrase clusters (sentences with the same meaning) to evaluate our tagging behavior and model output diversity improvement. Three English paraphrase cluster datasets were utilized, but no Chinese cluster datasets since we did not find such a dataset to be available.
One English and one Chinese paraphrase pair dataset were used.
Some of the datasets are widely used in paraphrase generation tasks, the others are adopted from other natural language tasks.
Below we briefly describe each dataset:

\smallskip \noindent \textbf{MS COCO Captions 2017 (MSCOCO)} \citep{lin2014microsoft} is an English dataset designed for automatic image caption generation. Five human-written caption descriptions were collected for each image. The total number of images is about 118k with 590k captions. 

\smallskip \noindent \textbf{Parabank} \citep{hu2019parabank} is an English sentence re-writing dataset. We took its eval dataset (ParabankEval), which contains 400 semantic paraphrase with an average of 14 paraphrases in each cluster. 

\smallskip \noindent \textbf{ComQA} is an English dataset of real user questions from the WikiAnswers community QA platform.
The questions are grouped into paraphrase clusters, of which 1,809 clusters have more than one question. 

\smallskip \noindent \textbf{Quora Question Pairs (QQP)} contain human-annotated duplicate English questions, with 50k training paraphrase pairs and 20k testing instances.

\smallskip \noindent \textbf{ATEC} is a Chinese dataset that comes from Ant Technology Exploration Conference Developer competition. It contains 14,946 financial question pairs that are semantically similar.

\medskip \noindent
For MSCOCO, ParabankEval, and ComQA, we randomly picked one sentence from each cluster as the paraphrase source sentence and the rest as the ground-truth references.
For the other datasets, we pick one sentence as the source and the other one as the reference.
We divided all datasets with 80\% in the training set and 20\% in the test set.

\begin{table*}[]
\centering
\begin{adjustbox}{max width=\textwidth}
\begin{tabular}{ccccccccccccccc}
\toprule
\multicolumn{1}{c}{Dataset} &  \multicolumn{4}{c}{MSCOCO} & \multicolumn{3}{c}{ComQA} & \multicolumn{3}{c}{ParabankEval} & \multicolumn{2}{c}{QQP} & \multicolumn{2}{c}{ATEC}      \\
\cmidrule(lr){1-1} \cmidrule(lr){2-5} \cmidrule(lr){6-8} \cmidrule(lr){9-11} \cmidrule(lr){12-13} \cmidrule(lr){14-15}
Tagger  & No Tag & Auto & NER  & Oracle & No Tag & NER  & Oracle & No Tag       & NER  & Oracle & No Tag & NER  & No Tag & NER  \\
\midrule
R       & 20.6   & 30.7 & 28.8 & 22.8   & 38.0     & 42.2 & 47.1   & 30.4         & 77.4 & 79.9   & 35.2   & 38.3 & 35.2   & 36.3 \\
R vs. S & 39.5   & 34.6 & 30.5 & 27.9   & 69.5   & 49.4 & 52.4   & 41.5         & 81.5 & 85.8   & 57.5   & 51.3 & 85.4   & 83.3 \\
T\%     &        & 99.7 & 99.3 & 99.9   &        & 91.3 & 92.7   &              & 99.1 & 100    &        & 80.2 &        & 96.1 \\
\bottomrule
\end{tabular}
\end{adjustbox}
\caption{Experiment results.\label{tab:all-results-small} Reading horizontally, the Tagger columns show the tagger (Auto Tagger, NER Tagger, Oracle Tagger, or no tagger) used during training for each dataset. Note that the Oracle Tagger is a realistic condition in this context because in a deployed application, the correct tags would be supplied by a human. The first row (``R'') refers to the result of 2-gram ROUGE scores of the generated paraphrases against references; this measures the {\it quality} of the paraphrases. The second row (``R vs.\ S'') refers to the 2-gram ROUGE scores of generated paraphrases against the source sentence; this measures the {\it diversity} of the paraphrases. In other words, the lower this value, the more diverse the paraphrases are. It is important to present both figures because paraphrase generation requires a balance between these two factors. The final row (``T\%'') shows the percentage of tagged substrings that remain unchanged during paraphrase generation; this demonstrates how well our model learns to preserve the content surrounded by tags. We can see that our tagging technique achieves the intended effect after fine-tuning.}
\end{table*}

\section{Results}

As a preface to our results, we emphasize that, to our knowledge, user-controllable paraphrasing in the manner that we have described is a novel capability.
That is, no previous work has addressed this problem---and thus, points of comparison are limited.
Further note that the point of our technique is {\it not} to establish state-of-the-art performance on these various datasets, but rather to illustrate our tagging feature. Nevertheless, it is worth noticing that our model outperforms the previous state-of-the-art paraphrase generation model \citep{lbow} on MSCOCO and QQP datasets in terms of ROUGE scores. 
We only present the results of ROUGE $2$-gram scores for brevity, but the results of other ROUGE scores are consistent.

We choose ROUGE~\citep{rouge} instead of BLEU~\citep{10.3115/1073083.1073135} as our evaluation metrics for both tagger performance and diversity evaluation because of the diversity loss that we introduced.
In an ideal dataset, the references of a source sentence should be able to cover all possible paraphrases.
In this case, precision-based metrics like BLEU should be an even more suitable evaluation metric of generated paraphrases.
However, the datasets we experimented on only contain a handful of references, far from covering all possibilities.
The substituted words and diversified descriptions encouraged by our minimized mutual information will all be false negatives since they are not covered by the reference, thus leading to low precision.

\subsection{Tagger Performance}

Experimental results and descriptions are shown in Table~\ref{tab:all-results-small}.
Compared to 118k semantic clusters in MSCOCO, ComQA and ParabankEval contain only 400 and 1809 clusters, respectively.
Therefore, only MSCOCO has enough data to train a BERT-based token classifier as its Auto Tagger, while ComQA and ParabankEval's Auto Tagger results are gibberish.

How should we read the results in Table~\ref{tab:all-results-small}?
We need to consider the ``R'' row and the ``R vs.\ S'' row together.
The values in the ``R'' row are 2-gram ROUGE scores; the higher the score, the more similar our generated paraphrase is to the reference sentences, and therefore the ``R'' score means how good our  paraphrases are---in other words, {\it quality}.
The values in the ``R vs.\ S'' row are 2-gram ROUGE scores of the generated paraphrase with respect to the source; the higher the score, the more similar our generated paraphrase is to the source sentence---in other words, {\it diversity}.
We desire both high quality and high diversity, in other words, a high ``R'' score and a low ``R vs.\ S'' score, but this is dataset dependent:\ if the references are similar to the sources (like in ParabankEval), both ``R'' scores and ``R vs.\ S'' scores will  inevitably be high; if the references are very different from the sources (like in the MSCOCO caption dataset), both ``R'' scores and ``R vs.\ S'' scores will tend to be low.

Why don't we use BLEU~\citep{10.3115/1073083.1073135} score instead? 
Consider again generating paraphrases for our running example:\ ``What are cheap lodging options in $\langle$tag$\rangle$ Beijing $\langle/$tag$\rangle$?''
We want our generated paraphrases to retain ``Beijing'' and change other parts of the sentence as much as possible (but preserving the semantic meaning).
Let’s suppose we have ground-truth references that contain all possible paraphrases.
If so, the paraphrase ``I’m visiting Beijing and trying to find cheap hotels.''\ will have a perfect ``R'' score and also a 0 ``R vs.\ S'' score (since it does not have any 2-gram overlaps with the source sentence); this will be the case with either ROUGE or BLEU.
For paraphrases like ``I’m visiting New York and trying to find cheap hotels.''\ or ``I’m visiting Beijing and trying to find the Great Wall.'', the ``R'' score will be lower.
A paraphrase like ``What are cheap hotels in Beijing?''\ will have a higher ``R vs.\ S'' score (lower diversity).
These are exactly the behaviors we desire.

More realistically, though, our references only contain a handful of paraphrases.
For example, what if the term ``cheap hotels'' never appears in our references? Then, even though we know that ``cheap hotels'' is a valid paraphrase of ``cheap lodging options'', the related 2-grams like ``cheap hotels'' and ``hotels in'' will all be false negatives. In this case, recall-based metrics like ROUGE will be more robust than precision-based metrics like BLEU.
Moreover, since we introduced the mutual information loss term to force our model to change the untagged parts of a sentence, it is likely that our model will generate correct paraphrases containing terms that  are not covered by the references, and those false negatives will result in artificially (and unfair) lower BLEU scores.
For this reason, we argue that ROUGE is the more appropriate metric in our study.

For three out of four English datasets, our proposed tagging approach yields not only better paraphrases (in terms of matching the reference) but more diverse paraphrases as well.
ATEC also yields similar behavior with the NER tagger.
However, ParabankEval seems to be an outlier here:\ the quality of the paraphrases increases dramatically (over double the score), although the generated output is far less diverse.
This is understandable since ParabankEval only has 400 semantic clusters in total and the sentences are usually two to three times longer than MSCOCO; the paraphrase generator did not see enough examples to generate diversified long paraphrases.

\subsection{Cross-Lingual Transfer}
\label{Language-Mixing}

Ideally, we desire a model with strong cross-lingual capabilities---for example, along the lines of previous work in tagging tasks~\cite{wu-dredze-2019-beto} and information retrieval~\cite{shi-etal-2020-cross}.
From a practical perspective, such capabilities can reduce the need for language-specific paraphrase training data.
From a scientific perspective, such explorations might help reveal language-agnostic ``universals'' for semantics.
In this section, we present two experiments that anecdotally provide some interesting observations.

In our first experiment, we fine-tuned the model only with Chinese sentence pairs.
During evaluation, we feed it English sentences and ask it to generate English paraphrases.
We do not provide a formal evaluation, but it appears that our model is able to generate English paraphrases fluently, including the ability to preserve tagged substrings.
For example, paraphrases of the input ``How do you get deleted $\langle$tag$\rangle$ Instagram $\langle/$tag$\rangle$ chats?'' include the following:
\begin{quote}
How do I get deleted  $\langle$tag$\rangle$ Instagram $\langle/$tag$\rangle$ messages?\\
How do I recover deleted  $\langle$tag$\rangle$ Instagram $\langle/$tag$\rangle$ messages?\\
How do you get deleted  $\langle$tag$\rangle$ Instagram $\langle/$tag$\rangle$ messages?\\
How do I get deleted  $\langle$tag$\rangle$ Instagram $\langle/$tag$\rangle$ posts?
\end{quote}

\noindent In the second experiment, we fine-tuned our model with English sentence pairs, and then feed it Chinese sentences and ask for Chinese paraphrases.
In this case, the model generates a mix of English and Chinese tokens, but, interestingly, the code switching occurs in a semantically coherent manner.
For example, paraphrases of sentence
\begin{CJK*}{UTF8}{gbsn}
``吃什么东西能$\langle$tag$\rangle$补肾$\langle/$tag$\rangle$呀?'' (translation, ``What foods can fortify the kidneys?''), where \textit{``补肾''} means ``fortify the kidneys''):
\begin{quote}
What is the best food to $\langle$tag$\rangle$补肾$\langle/$tag$\rangle$?\\
What are some foods that $\langle$tag$\rangle$补肾$\langle/$tag$\rangle$?\\
Which is the best food to $\langle$tag$\rangle$补肾$\langle/$tag$\rangle$?\\
What is the best thing to eat to $\langle$tag$\rangle$补肾$\langle/$tag$\rangle$?''
\end{quote}
When we paraphrase sentences without any tags in our second experiment, for example, ``手机如何快速散热?'' (translation, ``How to quickly dissipate the heat of a phone?'', where ``散热'' means ``dissipate heat'' and ``手机'' means ``phone''), the paraphrases are:
\begin{quote}
What is the best way to散热 your phone? \\
What is the best way to散热 your 手机?\\
What is the best way to散热 my phone?\\
What is the quickest way to散热 your phone?\\
What is the quickest way to散热 your 手机?
\end{quote}
\end{CJK*}

\noindent Based on the second experiment, it appears that tagged Chinese tokens are in general preserved.
In the absence of tags, the model is mostly performing translation, as most of the generated tokens are in English (as well as overall word order and grammar).
However, as the above examples show, some Chinese tokens are idiosyncratically preserved.
Most interestingly, the code switches are semantically coherent.

Although these results are at best anecdotal, it rules out obvious and low-hanging fruit in cross-lingual transfer capabilities.
We suspect these observations point to the dominance of English in the pretraining of mBART---it seems like the case that multi-lingual capabilities are pivoting through English.
Even when trained on Chinese paraphrase pairs, results suggest that they are likely mapped to English latent semantic space, and that Chinese is easily ``forgotten''.

\section{Conclusions and Future Work}

This paper tackles a practical, real-world problem in paraphrase generation that to our knowledge has not been previously addressed:\ there are tokens that a user might wish to preserve verbatim for a variety of reasons.
We further assume, that in many scenarios, the user knows exactly what those tokens are.
This leads to our relatively straightforward solution---we provide the user with tags whose semantics are ``don't change me''.

The contribution of our work is a language-agnostic implementation of this capability using a pretrained sequence-to-sequence model (mBART), coupled with an objective that encourages diversity in the parts of the input that are not surrounded by tags.
Evaluations on both English and Chinese paraphrase datasets demonstrate the empirical success of our proposed model, and additional experiments reveal interesting observations about cross-lingual transfer effects, potentially paving the way for future studies.

\bibliography{eacl2021}

\begin{thebibliography}{10}
\expandafter\ifx\csname natexlab\endcsname\relax\def\natexlab#1{#1}\fi

\bibitem[{Chen et~al.(2015)Chen, Fang, Lin, Vedantam, Gupta, Doll{\'a}r, and
  Zitnick}]{lin2014microsoft}
Xinlei Chen, Hao Fang, Tsung-Yi Lin, Ramakrishna Vedantam, Saurabh Gupta, Piotr
  Doll{\'a}r, and C.~Lawrence Zitnick. 2015.
\newblock Microsoft {COCO} captions: Data collection and evaluation server.
\newblock \emph{arXiv preprint arXiv:1504.00325}.

\bibitem[{Devlin et~al.(2019)Devlin, Chang, Lee, and
  Toutanova}]{devlin2018bert}
Jacob Devlin, Ming-Wei Chang, Kenton Lee, and Kristina Toutanova. 2019.
\newblock {BERT}: Pre-training of deep bidirectional transformers for language
  understanding.
\newblock In \emph{Proceedings of the 2019 Conference of the North {A}merican
  Chapter of the Association for Computational Linguistics: Human Language
  Technologies, Volume 1 (Long and Short Papers)}, pages 4171--4186.

\bibitem[{Fu et~al.(2019)Fu, Feng, and Cunningham}]{lbow}
Yao Fu, Yansong Feng, and John~P. Cunningham. 2019.
\newblock Paraphrase generation with latent bag of words.
\newblock In \emph{Advances in Neural Information Processing Systems}.

\bibitem[{Hu et~al.(2019)Hu, Rudinger, Post, and Van~Durme}]{hu2019parabank}
J.~Edward Hu, Rachel Rudinger, Matt Post, and Benjamin Van~Durme. 2019.
\newblock {ParaBank}: Monolingual bitext generation and sentential paraphrasing
  via lexically-constrained neural machine translation.
\newblock In \emph{Proceedings of the AAAI Conference on Artificial
  Intelligence}, volume~33, pages 6521--6528.

\bibitem[{Lin(2004)}]{rouge}
Chin-Yew Lin. 2004.
\newblock {ROUGE}: A package for automatic evaluation of summaries.
\newblock In \emph{Text Summarization Branches Out}, pages 74--81, Barcelona,
  Spain.

\bibitem[{Liu et~al.(2020)Liu, Gu, Goyal, Li, Edunov, Ghazvininejad, Lewis, and
  Zettlemoyer}]{mbart}
Yinhan Liu, Jiatao Gu, Naman Goyal, Xian Li, Sergey Edunov, Marjan
  Ghazvininejad, Mike Lewis, and Luke Zettlemoyer. 2020.
\newblock Multilingual denoising pre-training for neural machine translation.
\newblock \emph{arXiv preprint arXiv:2001.08210}.

\bibitem[{Papineni et~al.(2002)Papineni, Roukos, Ward, and
  Zhu}]{10.3115/1073083.1073135}
Kishore Papineni, Salim Roukos, Todd Ward, and Wei-Jing Zhu. 2002.
\newblock {BLEU}: A method for automatic evaluation of machine translation.
\newblock In \emph{Proceedings of the 40th Annual Meeting on Association for
  Computational Linguistics}, ACL '02, page 311–318.

\bibitem[{Shi et~al.(2020)Shi, Bai, and Lin}]{shi-etal-2020-cross}
Peng Shi, He~Bai, and Jimmy Lin. 2020.
\newblock Cross-lingual training of neural models for document ranking.
\newblock In \emph{Findings of the Association for Computational Linguistics:
  EMNLP 2020}, pages 2768--2773.

\bibitem[{Strubell et~al.(2017)Strubell, Verga, Belanger, and
  McCallum}]{strubell2017fast}
Emma Strubell, Patrick Verga, David Belanger, and Andrew McCallum. 2017.
\newblock Fast and accurate entity recognition with iterated dilated
  convolutions.
\newblock \emph{arXiv preprint arXiv:1702.02098}.

\bibitem[{Wu and Dredze(2019)}]{wu-dredze-2019-beto}
Shijie Wu and Mark Dredze. 2019.
\newblock Beto, bentz, becas: The surprising cross-lingual effectiveness of
  {BERT}.
\newblock In \emph{Proceedings of the 2019 Conference on Empirical Methods in
  Natural Language Processing and the 9th International Joint Conference on
  Natural Language Processing (EMNLP-IJCNLP)}, pages 833--844, Hong Kong,
  China.

\end{thebibliography}
\bibliographystyle{acl_natbib}

\end{document}